\PassOptionsToPackage{usenames,dvipsnames,svgnames}{xcolor}
\PassOptionsToPackage{hyphens}{url}
\documentclass[11pt,a4paper,final]{article}
\usepackage[hyperref]{acl2020}
\usepackage{times}
\usepackage{latexsym}
\usepackage{polozov_paper}

\aclfinalcopy %
\def\aclpaperid{2884} %

\allowdisplaybreaks

\newcommand\bestresult{57.2\%}
\newcommand\bestresultwithbert{65.6\%}

\title{{RAT-SQL}: Relation-Aware Schema Encoding and Linking \\ for Text-to-{SQL} Parsers}

\author{
    Bailin Wang\Thanks{~Equal contribution. Order decided by a coin toss.}\,\ \Thanks{~Work done during an internship at Microsoft Research.}\\
    University of Edinburgh \\
    \email{bailin.wang@ed.ac.uk} \\\And
    Richard Shin\footnotemark[1]\,\ \Thanks{~Work done partly affiliated with Microsoft Research. Now at Microsoft: \protect\email{richard.shin@microsoft.com}.} \\
    UC Berkeley \\
    \email{ricshin@cs.berkeley.edu} \\\AND
    Xiaodong Liu \quad Oleksandr Polozov \quad Matthew Richardson \\
    Microsoft Research, Redmond \\
    \email{{xiaodl,polozov,mattri}@microsoft.com}}

\begin{document}
\newcommand{\question}{Q}
\newcommand{\word}{q}
\newcommand{\sColumnSet}{\mathcal{C}}
\newcommand{\sColumn}{c}
\newcommand{\sTableSet}{\mathcal{T}}
\newcommand{\sTable}{t}
\newcommand{\sql}{P}
\newcommand{\type}{\tau}
\newcommand{\schema}{\mathcal{S}}
\renewcommand{\ast}{T}
\newcommand{\hidden}{\vect{h}}
\newcommand{\action}{a}
\newcommand{\encoder}{f_{\text{enc}}}
\newcommand{\decoder}{f_{\text{dec}}}
\newcommand{\sGraph}{\mathcal{G}}
\newcommand{\inputGraph}{\sGraph_{\question}}

\maketitle

\begin{abstract}
    When translating natural language questions into SQL queries to answer questions from a database,
    contemporary semantic parsing models struggle to generalize to unseen database schemas.
    The generalization challenge lies in (a) encoding the database relations in an accessible way for the semantic
    parser, and (b) modeling alignment between database columns and their mentions in a given query.
    We present a unified framework, based on the relation-aware self-attention mechanism, to address schema encoding,
    schema linking, and feature representation within a text-to-SQL encoder.
    On the challenging Spider dataset this framework boosts the exact match accuracy to {\bestresult}, surpassing its best counterparts by 8.7\% absolute improvement. 
    Further augmented with BERT, it achieves the new state-of-the-art performance of {\bestresultwithbert} on the Spider leaderboard.
    In addition, we observe qualitative improvements in the model's understanding of schema linking and alignment.
    Our implementation will be open-sourced at \mbox{\url{https://github.com/Microsoft/rat-sql}}.
\end{abstract}

\section{Introduction}
\label{sec:introduction}
The ability to effectively query databases with natural language (NL) unlocks the power of large datasets
to the vast majority of users who are not proficient in query languages.
As such, a large body of research has focused on the task of translating NL questions into SQL queries
that existing database software can execute.

\begin{figure*}
    \centering
    \includegraphics[width=\textwidth]{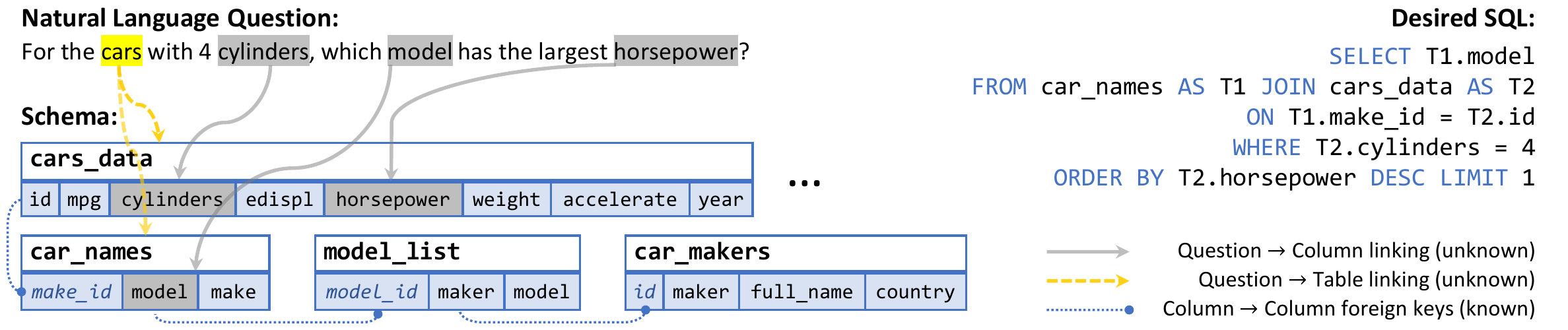}
    \caption{
        A challenging text-to-SQL task from the Spider dataset.
    }%
    \label{fig:intro:example}
    \vspace{-\baselineskip}
\end{figure*}

The development of large annotated datasets of questions and the corresponding SQL queries has catalyzed
progress in the field.
In contrast to prior semantic parsing datasets~\citep{finegan-dollakImprovingTexttoSQLEvaluation2018}, new tasks such as
WikiSQL~\citep{zhongSeq2SQLGeneratingStructured2017} and Spider~\citep{data-spider} pose the real-life challenge of
\emph{generalization to unseen database schemas}.
Every query is conditioned on a multi-table database schema, and the databases do not overlap between the train and test
sets.

Schema generalization is challenging for three interconnected reasons.
First, any text-to-SQL parsing model must encode the schema into
representations suitable for decoding a SQL query that might involve the given columns or tables.
Second, these representations should encode all the information about the schema such as its column types, foreign
key relations, and primary keys used for database joins.
Finally, the model must recognize NL used to refer to columns and tables, which might differ from
the referential language seen in training.
The latter challenge is known as \emph{schema linking} -- aligning entity references in the question to the
intended schema columns or tables.

While the question of \emph{schema encoding} has been studied in recent literature~\citep{gnnsql}, \emph{schema linking}
has been relatively less explored.
Consider the example in \Cref{fig:intro:example}.
It illustrates the challenge of ambiguity in linking:
while \emph{``model''} in the question refers to \texttt{car\_names.model} rather than \texttt{model\_list.model},
\emph{``cars''} actually refers to both \texttt{cars\_data} and \texttt{car\_names} (but not \texttt{car\_makers}) for
the purpose of table joining.
To resolve the column/table references properly, the semantic parser must take into account both the known schema
relations (\eg foreign keys) and the question context.

Prior work~\citep{gnnsql} addressed the schema representation problem by encoding the directed graph of foreign key
relations in the schema with a graph neural network (GNN).
While effective, this approach has two important shortcomings.
First, it does not contextualize schema encoding with the question, thus making reasoning
about schema linking difficult after both the column representations and question word representations are built.
Second, it limits information propagation during schema encoding to the predefined graph of foreign key relations.
The advent of self-attentional mechanisms in NLP~\citep{vaswaniAttentionAllYou2017} shows that
global reasoning is crucial to effective representations of relational structures.
However, we would like any global reasoning to still take into account the aforementioned schema relations.

In this work, we present a unified framework, called RAT-SQL,%
\footnote{\underline{\bfseries R}elation-\underline{\bfseries A}ware \underline{\bfseries T}ransformer.}
for encoding relational structure in the database schema and a given question.
It uses \emph{relation-aware self-attention} to combine global reasoning over the schema entities and question words
with structured reasoning over predefined schema relations.
We then apply RAT-SQL to the problems of schema encoding and schema linking.
As a result, we obtain \bestresult{} exact match accuracy on the Spider test set.
At the time of writing, this result is the state of the art among models unaugmented with pretrained BERT
embeddings~--~and further reaches to the overall state of the art (\bestresultwithbert) when RAT-SQL is augmented with
BERT.
In addition, we experimentally demonstrate that RAT-SQL enables the model to build more accurate internal
representations of the question's true alignment with schema columns and tables.

\section{Related Work}
\label{sec:related-work}

Semantic parsing of NL to SQL recently surged in popularity thanks to the creation of
two new multi-table datasets with the challenge of schema generalization --
WikiSQL~\citep{zhongSeq2SQLGeneratingStructured2017} and Spider~\citep{data-spider}.
Schema encoding is not as challenging in WikiSQL as in Spider because it lacks multi-table relations.
Schema linking is relevant for both tasks but also more challenging in Spider due to the richer NL
expressiveness and less restricted SQL grammar observed in it.
The state of the art semantic parser on WikiSQL~\citep{xsql} achieves a test set accuracy of 91.8\%, significantly
higher than the state of the art on Spider.

The recent state-of-the-art models evaluated on Spider use various attentional architectures for question/schema
encoding and AST-based structural architectures for query decoding.
IRNet~\citep{irnet} encodes the question and schema separately with LSTM and self-attention respectively, augmenting
them with custom type vectors for schema linking.
They further use the AST-based decoder of \citet{yinSyntacticNeuralModel2017a} to decode a query in an intermediate
representation (IR) that exhibits higher-level abstractions  than SQL.
\citet{gnnsql} encode the schema with a GNN and a similar grammar-based decoder.
Both works emphasize schema encoding and schema linking, but design separate featurization
techniques to augment \emph{word vectors} (as opposed to \emph{relations between words and columns}) to resolve it.
In contrast, the RAT-SQL framework provides a unified way to encode arbitrary relational information among inputs.

Concurrently with this work, \citet{bogin2019global} published Global-GNN, a different approach to schema linking for
Spider, which applies global reasoning between question words and schema columns/tables.
Global reasoning is implemented by \emph{gating} the GNN that encodes the schema using the question token representations.
This differs from RAT-SQL in two important ways:
(a) question word representations influence the schema representations but not vice versa, and
(b) like in other GNN-based encoders, message propagation is limited to the schema-induced edges such as
foreign key relations.
In contrast, our relation-aware transformer mechanism allows encoding arbitrary relations between question words and
schema elements explicitly, and these representations are computed jointly over all inputs using self-attention.

We use the same formulation of relation-aware self-attention as \citet{shawSelfAttentionRelativePosition2018}.
However, they only apply it to sequences of words in the context of machine translation, and as such, their relation types only encode the relative distance between two words.
We extend their work and show that relation-aware self-attention can effectively encode more complex relationships within an unordered set of elements (in our case, columns and tables within a database schema as well as relations between the schema and the question).
To the best of our knowledge, this is the first application of relation-aware self-attention to joint representation learning with both predefined and softly induced relations in the input structure.
\citet{Hellendoorn2020Global} develop a similar model concurrently with this work, where they use relation-aware self-attention to encode data flow structure in source code embeddings.

\citet{sun-etal-2018-open} use a heterogeneous graph of KB facts and relevant documents for open-domain question answering.
The nodes of their graph are analogous to the database schema nodes in RAT-SQL, but RAT-SQL also incorporates the question in the same formalism to enable joint representation learning between the question and the schema.

\section{Relation-Aware Self-Attention}%
\label{sec:rat}

First, we introduce \emph{relation-aware self-attention}, a model for embedding semi-structured input sequences in a way
that jointly encodes pre-existing relational structure in the input as well as induced ``soft'' relations between
sequence elements in the same embedding.
Our solutions to schema embedding and linking naturally arise as features implemented in this framework.

Consider a set of inputs $ X = \left\{ \vect{x_i} \right\}_{i=1}^n $ where $\vect{x_i} \in \Reals^{d_x}$.
In general, we consider it an unordered set, although $\vect{x_i}$ may be imbued with positional embeddings to add an
explicit ordering relation.
A \emph{self-attention} encoder, or \emph{Transformer}, introduced by \citet{vaswaniAttentionAllYou2017}, is a stack of
\emph{self-attention layers} where each layer (consisting of $H$ \emph{heads}) transforms each $\vect{x_i}$ into
$\vect{y_i} \in \Reals^{d_x}$ as follows:
\begin{gather}
    \small
    \begin{aligned}
        e_{ij}^{(h)} &= \frac{\vect{x_i} W_Q^{(h)} (\vect{x_j} W_K^{(h)})^\top}{\sqrt{d_z / H}}
        ;\quad
        \alpha_{ij}^{(h)} = \softmax_{j} \bigl\{ e_{ij}^{(h)} \bigr\}
        \\
        \vect{z}_i^{(h)} &= \sum_{j=1}^n \alpha_{ij}^{(h)} (\vect{x_j} W_V^{(h)})
        ;\quad
        \vect{z}_i = \text{Concat}\bigl(\vect{z}_i^{(1)}, \cdots, \vect{z}_i^{(H)}\bigr)
        \\
        \vect{\tilde{y}_i} &= \text{LayerNorm}(\vect{x_i} + \vect{z}_i)
        \\
        \vect{y_i} &= \text{LayerNorm}(\vect{\tilde{y}_i} + \text{FC}(\text{ReLU}(\text{FC}(\vect{\tilde{y}_i})))
    \end{aligned}
    \raisetag{1.1\baselineskip}
    \label{eq:transformer}
\end{gather}
where FC is a fully-connected layer, LayerNorm is \emph{layer normalization}~\citep{baLayerNormalization2016},
\mbox{$1 \le h \le H$}, and \mbox{$W_Q^{(h)}, W_K^{(h)}, W_V^{(h)} \in \mathbb{R}^{d_x \times (d_x / H)}$}.

One interpretation of the embeddings computed by a Transformer is that each head of each layer computes a
\emph{learned relation} between all the input elements $\vect{x_i}$, and the strength of this relation is encoded in the
attention weights $\alpha_{ij}^{(h)}$.
However, in many applications (including text-to-SQL parsing) we are aware of some preexisting relational
features between the inputs, and would like to bias our encoder model toward them.
This is straightforward for non-relational features (represented directly in each $\vect{x_i}$).
We could limit the attention computation only to the ``hard'' edges where the preexisting relations are known to hold.
This would make the model similar to a \emph{graph attention network}~\citep{velickovic2018graph}, and would also
impede the Transformer's ability to learn \emph{new} relations.
Instead, RAT provides a way to communicate known relations to the encoder by adding their representations
to the attention mechanism.

\citet{shawSelfAttentionRelativePosition2018} describe a way to represent \emph{relative position information} in a
self-attention layer by changing \Cref{eq:transformer} as follows:
\begin{equation}
    \label{eq:transformer:self}
    \begin{aligned}
        e_{ij}^{(h)} &= \frac{\vect{x_i} W_Q^{(h)} (\vect{x_j} W_K^{(h)} + \vect{\color{red} r_{ij}^K})^\top}{\sqrt{d_z
        / H}} \\
        \vect{z}_i^{(h)} &= \sum_{j=1}^n \alpha_{ij}^{(h)} (\vect{x_j} W_V^{(h)} + \vect{\color{red} r_{ij}^V}).
    \end{aligned}
\end{equation}
Here the $\vect{r_{ij}}$ terms encode the known relationship between the two elements $\vect{x_i}$ and $\vect{x_j}$ in
the input.
While \citeauthor{shawSelfAttentionRelativePosition2018} used it exclusively for relative position representation, we
show how to use the same framework to effectively bias the Transformer toward arbitrary relational information.

Consider $R$ relational features, each a binary relation
\mbox{$ \mathcal{R}^{(s)} \subseteq X \times X $} \mbox{$(1 \le s \le R)$}.
The RAT framework represents all the pre-existing features for each edge $(i,j)$ as $\vect{r_{ij}^K} = \vect{r_{ij}^V} =
\mathrm{Concat}\bigl(\vect{\rho^{(1)}_{ij}}, \dots, \vect{\rho^{(R)}_{ij}}\bigr)$ where
each $\vect{\rho^{(s)}_{ij}}$ is either a \emph{learned embedding} for the relation $ \mathcal{R}^{(s)} $ if the
relation holds for the corresponding edge (\ie if $ (i, j) \in \mathcal{R}^{(s)}$), or a zero vector of appropriate
size.
In the following section, we will describe the set of relations our RAT-SQL model uses to encode a given database schema.

\section{RAT-SQL}
\label{sec:our-encoder}
We now describe the RAT-SQL framework and its application to the problems of schema encoding and linking.
First, we formally define the text-to-SQL semantic parsing problem and its components.
In the rest of the section, we present our implementation of schema linking in the RAT framework.

\begin{table*}[t]
    \centering
    \begin{tabular}{lllp{7cm}}
        \toprule
        Type of $x$ & Type of $y$ & Edge label & Description \\
        \midrule
        \multirow{3}{*}{Column} & \multirow{3}{*}{Column}
   & \textsc{Same-Table}    & $x$ and $y$ belong to the same table. \\
 & & \textsc{Foreign-Key-Col-F} & $x$ is a foreign key for $y$. \\
 & & \textsc{Foreign-Key-Col-R} & $y$ is a foreign key for $x$. \\
 \midrule
 \multirow{2}{*}{Column} & \multirow{2}{*}{Table}
   & \textsc{Primary-Key-F}   & $x$ is the primary key of $y$. \\
 & & \textsc{Belongs-To-F}    & $x$ is a column of $y$ (but not the primary key). \\
 \midrule
 \multirow{2}{*}{Table} & \multirow{2}{*}{Column}
   & \textsc{Primary-Key-R}   & $y$ is the primary key of $x$. \\
 & & \textsc{Belongs-To-R}    & $y$ is a column of $x$ (but not the primary key). \\
 \midrule
 \multirow{3}{*}{Table} & \multirow{3}{*}{Table}
   & \textsc{Foreign-Key-Tab-F}   & Table $x$ has a foreign key column in $y$. \\
 & & \textsc{Foreign-Key-Tab-R}   & Same as above, but $x$ and $y$ are reversed. \\
 & & \textsc{Foreign-Key-Tab-B}   & $x$ and $y$ have foreign keys in both directions. \\
 \bottomrule
\end{tabular}
\caption{Description of edge types present in the directed graph $\sGraph$ created to represent the schema.
    An edge exists from source node $x \in \schema$ to target node $y \in \schema$ if the pair fulfills one of the
    descriptions listed in the table, with the corresponding label.
Otherwise, no edge exists from $x$ to $y$.}
\label{table:schema-graph-edges}
\vspace{-1\baselineskip}
\end{table*}

\subsection{Problem Definition}
Given a natural language question $\question$ and a schema $\schema = \langle \sColumnSet, \sTableSet \rangle$ for a
relational database, our goal is to generate the corresponding SQL $\sql$.
Here the question $\question = \word_1 \dots \word_{|\question|}$ is a sequence of words,
and the schema consists of columns $\sColumnSet = \{ \sColumn_1, \dots, \sColumn_{|\sColumnSet|} \}$
and tables $\sTableSet = \left\{ \sTable_1, \dots, \sTable_{|\sTableSet|} \right\}$.
Each column name~$\sColumn_i$ contains words $\sColumn_{i,1}, \dots, \sColumn_{i, |\sColumn_i|}$ and
each table name~$\sTable_i$ contains words $\sTable_{i,1}, \dots, \sTable_{i, |\sTable_i|}$.
The desired program $\sql$ is represented as an \emph{abstract syntax tree} $\ast$ in the context-free grammar of SQL.

Some columns in the schema are \emph{primary keys}, used for uniquely indexing the corresponding table,
and some are \emph{foreign keys}, used to reference a primary key column in a different table.
In addition, each column has a \emph{type} $\type \in \{ \texttt{number},\, \texttt{text} \}$.

\begin{figure}[t]
    \centering
    \includegraphics[width=\columnwidth]{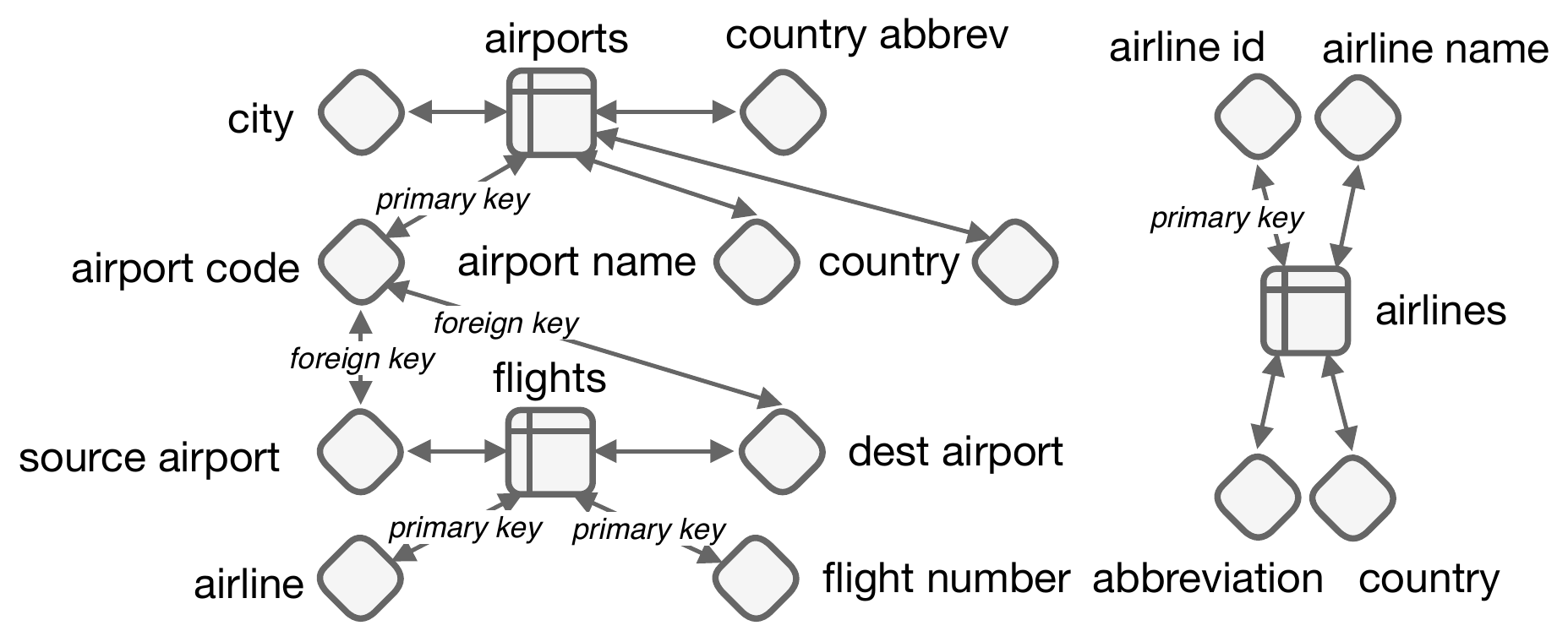}
    \vspace{-1\baselineskip}
    \caption{An illustration of an example schema as a graph $\sGraph$. We do not depict all the edges and label types
        of \cref{table:schema-graph-edges} to reduce clutter.}
    \vspace{-1\baselineskip}
    \label{fig:schema-graph}
\end{figure}

Formally, we represent the database schema as a directed
graph~$\sGraph = \langle \mathcal{V}, \mathcal{E} \rangle $.
Its nodes $ \mathcal{V} = \sColumnSet \cup \sTableSet $ are the columns and tables of the schema, each labeled with the
words in its name (for columns, we prepend their type $\type$ to the label).
Its edges $ \mathcal{E} $ are defined by the pre-existing database relations, described in
\Cref{table:schema-graph-edges}.
\Cref{fig:schema-graph} illustrates an example graph (with a subset of actual edges and labels).

While $ \sGraph $ holds all the known information about the schema, it is insufficient for appropriately encoding a
previously unseen schema \emph{in the context of the question $\question$}.
We would like our representations of the schema $\schema$ and the question $\question$ to be \emph{joint}, in particular
for modeling the alignment between them.
Thus, we also define the \emph{question-contextualized schema graph}
$ \inputGraph = \langle \mathcal{V}_{\question}, \mathcal{E}_{\question} \rangle $
where $ \mathcal{V}_{\question} = \mathcal{V} \cup \question = \sColumnSet \cup \sTableSet \cup \question$ includes
nodes for the question words (each labeled with a corresponding word), and
$ \mathcal{E}_{\question} = \mathcal{E} \cup \mathcal{E}_{\question \leftrightarrow \schema} $
are the schema edges $ \mathcal{E} $ extended with additional special relations between the question words and schema
members, detailed in the rest of this section.

For modeling text-to-SQL generation, we adopt the \emph{encoder-decoder framework}.
Given the input as a graph~$\inputGraph$, the \emph{encoder} $\encoder$ embeds it into joint representations
$\vect{c}_{i}$, $\vect{t}_{i}$, $\vect{q}_{i}$ for each column $\sColumn_i \in \sColumnSet$, table $\sTable_i \in
\sTableSet$, and question word $\word \in \question$ respectively.
The decoder $\decoder$ then uses them to compute a distribution $ \Pr( \sql \mid \inputGraph ) $ over
the SQL programs.

\subsection{Relation-Aware Input Encoding}
\label{sec:rel-attn}

\begin{figure}
    \centering
    \includegraphics[width=0.9\columnwidth]{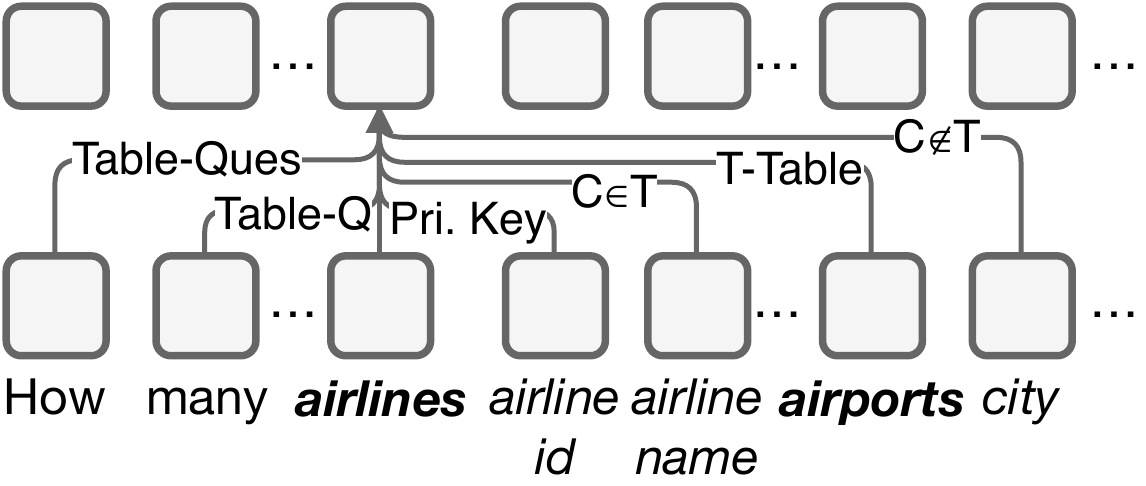}
    \caption{One RAT layer in the schema encoder.}
    \label{fig:rel-attn}
    \vspace{-1\baselineskip}
\end{figure}

Following the state-of-the-art NLP literature, our encoder first obtains the \emph{initial}
representations $\vect{c}_{i}^\text{init}$, $\vect{t}_{i}^\text{init}$
for every node of $\sGraph$ by
\textbf{(a)} retrieving a pre-trained Glove embedding ~\citep{pennington2014glove} for each word, and
\textbf{(b)} processing the embeddings in each multi-word label with a bidirectional LSTM (\mbox{BiLSTM})~\citep{lstm}.
It also runs a separate BiLSTM over the question $\question$ to obtain initial word representations
$\vect{q}_{i}^\text{init}$.

The initial representations $\vect{c}_{i}^\text{init}$, $\vect{t}_{i}^\text{init}$, and $\vect{q}_{i}^\text{init}$
are independent of each other and devoid of any relational information known to hold in $ \mathcal{E}_{\question} $.
To produce joint representations for the entire input graph $\inputGraph$, we use the relation-aware self-attention
mechanism (\Cref{sec:rat}).
Its input $X$ is the set of all the node representations in $\inputGraph$:
\[
    X = (\vect{c}_{1}^\text{init}, \cdots, \vect{c}_{|\sColumnSet|}^\text{init},
         \vect{t}_{1}^\text{init}, \cdots, \vect{t}_{|\sTableSet|}^\text{init},
         \vect{\word}_{1}^\text{init}, \cdots, \vect{\word}_{|\question|}^\text{init}).
\]
The encoder $\encoder$ applies a stack of $N$ relation-aware self-attention layers to $X$,
with separate weight matrices in each layer.
The final representations $\vect{c}_{i}$, $\vect{t}_{i}$, $\vect{q}_{i}$ produced by the $N^{\text{th}}$ layer
constitute the output of the whole encoder.

Alternatively, we also consider pre-trained BERT~\citep{bert} embeddings to obtain the initial representations.
Following \cite{huangMusicTransformer2018,EditSQL2019}, we feed $X$ to the BERT and use the last hidden states as the initial representations before proceeding with the RAT layers.\footnote{In this case, the initial representations $\vect{c}_{i}^\text{init}$, $\vect{t}_{i}^\text{init}$, $\vect{q}_{i}^\text{init}$ are not strictly independent although still yet uninfluenced by $\mathcal{E}$.}

Importantly, as detailed in \Cref{sec:rat}, every RAT layer uses self-attention between \emph{all elements of the input
graph $\inputGraph$} to compute new contextual representations of question words and schema members.
However, this self-attention is \emph{biased} toward some pre-defined relations using the edge vectors
$\vect{r_{ij}^K}, \vect{r_{ij}^V}$ in each layer.
We define the set of used relation types in a way that directly addresses the challenges of schema embedding and
linking.
Occurrences of these relations between the question and the schema constitute the edges $ \mathcal{E}_{\question
\leftrightarrow \schema} $.
Most of these relation types address schema linking (\Cref{sec:schema-linking});
we also add some auxiliary edges to aid schema encoding
(see Appendix \ref{app:schema-encoding}).

\subsection{Schema Linking}
\label{sec:schema-linking}
Schema linking relations in $ \mathcal{E}_{\question \leftrightarrow \schema} $ aid the model with aligning column/table
references in the question to the corresponding schema columns/tables.
This alignment is implicitly defined by two kinds of information in the input: \emph{matching names} and \emph{matching
values}, which we detail in order below.

\paragraph{Name-Based Linking}
Name-based linking refers to \emph{exact} or \emph{partial} occurrences of the column/table names in the question,
such as the occurrences of \emph{``cylinders''} and \emph{``cars''} in the question in \Cref{fig:intro:example}.
Textual matches are the most explicit evidence of question-schema alignment and as such, one might expect them to be
directly beneficial to the encoder.
However, in all our experiments the representations produced by vanilla self-attention were insensitive to
textual matches even though their initial representations were identical.
\citet{brunner2019identifiability} suggest that representations produced by Transformers mix the information from
different positions and cease to be directly interpretable after 2+ layers, which might explain our observations.
Thus, to remedy this phenomenon, we explicitly encode name-based linking using RAT relations.

Specifically, for all n-grams of length 1 to 5 in the question,
we determine (1) whether it exactly matches the name of a column/table (\emph{exact match});
or (2) whether the n-gram is a subsequence of the name of a column/table (\emph{partial match}).\footnote{This procedure
matches that of \citet{irnet}, but we use the matching information differently in RAT.}
Then, for every $(i,j)$ where $x_i \in \question$, $x_j \in \schema$ (or vice versa), we set
$r_{ij} \in  \mathcal{E}_{\question \leftrightarrow \schema} $ to
\textsc{Question-Column}-\textsf{M}, \textsc{Question-Table}-\textsf{M}, \textsc{Column-Question}-\textsf{M} or
\textsc{Table-Question}-\textsf{M} depending on the type of $x_i$ and $x_j$.
Here $\textsf{M}$ is one of \textsc{ExactMatch}, \textsc{PartialMatch}, or \textsc{NoMatch}.

\paragraph{Value-Based Linking}
Question-schema alignment also occurs when the question mentions any \emph{values} that occur in the database and
consequently participate in the desired SQL, such as \emph{``4''} in \Cref{fig:intro:example}.
While this example makes the alignment explicit by mentioning the column name \emph{``cylinders''}, many real-world
questions do not.
Thus, linking a value to the corresponding column requires background knowledge.

The database itself is the most comprehensive and readily available source of knowledge about possible values, but also
the most challenging to process in an end-to-end model because of the privacy and speed impact.
However, the RAT framework allows us to outsource this processing to the database engine to \emph{augment} $\inputGraph$
with potential value-based linking without exposing the model itself to the data.
Specifically, we add a new \textsc{Column-Value} relation between any word $\word_i$ and column name $\sColumn_j$ s.t.
$\word_i$ occurs as a value (or a full word within a value) of $\sColumn_j$.
This simple approach drastically improves the performance of RAT-SQL (see \Cref{sec:experiments}).
It also directly addresses the aforementioned DB challenges:
\textbf{(a)} the model is never exposed to database content that does not occur in the question,
\textbf{(b)} word matches are retrieved quickly via DB indices \& textual search.

\paragraph{Memory-Schema Alignment Matrix}
Our intuition suggests that the columns and tables which occur in the SQL $\sql$ will generally have a corresponding reference in the natural language question.
To capture this intuition in the model, we apply relation-aware attention as a pointer mechanism between every memory element in $y$ and all the columns/tables to compute explicit \emph{alignment matrices} $L^\text{col} \in \mathbb{R}^{|y| \times |\sColumnSet|}$ and $L^\text{tab} \in \mathbb{R}^{|y| \times |\sTableSet|}$:
\begin{align}
\label{eq:align_mat}
    \tilde{L}^\text{col}_{i,j} &= \frac{y_i W_Q^\text{col} (\vect{c}^\text{final}_j W_K^\text{col} +
    \vect{r}_{ij}^K)^\top}{\sqrt{d_x}} \\
    \tilde{L}^\text{tab}_{i,j} &= \frac{y_i W_Q^\text{tab} (\vect{t}^\text{final}_j W_K^\text{tab} +
    \vect{r}_{ij}^K)^\top}{\sqrt{d_x}} \notag\\
    L^\text{col}_{i,j} &= \softmax_{j} \bigl\{ \tilde{L}^{\text{col}}_{i,j} \bigr\} \qquad
    L^\text{tab}_{i,j} = \softmax_{j} \bigl\{ \tilde{L}^{\text{tab}}_{i,j} \bigr\} \notag
\end{align}

Intuitively, the alignment matrices in \cref{eq:align_mat} should resemble the real discrete alignments, therefore should respect certain constraints like sparsity.
When the encoder is sufficiently parameterized, sparsity tends to arise with learning, but we can also encourage it with an explicit objective.
\Cref{app:align_loss} presents this objective and discusses our experiments with sparse alignment in RAT-SQL.

\subsection{Decoder}
\label{sec:decoder}
The decoder $\decoder$ of RAT-SQL follows the tree-structured architecture of \citet{yinSyntacticNeuralModel2017a}.
It generates the SQL $\sql$ as an abstract syntax tree in depth-first traversal order,
by using an LSTM to output a sequence of \emph{decoder actions} that either (i) expand the last generated node into a grammar rule, called \textsc{ApplyRule}; or when completing a leaf node, (ii) choose a
column/table from the schema, called \textsc{SelectColumn} and \textsc{SelectTable}.

Formally,
\(
    \Pr(\sql \mid \mathcal{Y}) = \prod_t \Pr(a_t \mid a_{<t},\, \mathcal{Y})
\)
where $\mathcal{Y} = \encoder(\inputGraph)$ is the final encoding of the question and schema,
and $a_{<t}$ are all the previous actions.
In a tree-structured decoder, the LSTM state is updated as
$\vect{m}_t, \vect{h}_t  = f_{\text{LSTM}}\left( [\vect{a}_{t-1} \conc \vect{z}_t \conc \vect{h}_{p_t} \conc
\vect{a}_{p_t} \conc \vect{n}_{f_t}],\ \vect{m}_{t-1}, \vect{h}_{t-1} \right)$
where $\vect{m}_t$ is the LSTM cell state, $\vect{h}_t$ is the LSTM output at step $t$, $\vect{a}_{t-1}$ is the
embedding of the previous action, $p_t$ is the step corresponding to expanding the parent AST node of the current node,
and $\vect{n}_{f_t}$ is the embedding of the current node type.
Finally, $\vect{z}_t$ is the context representation, computed using multi-head attention (with 8 heads) on
$\vect{h}_{t-1}$ over $\mathcal{Y}$.

For \textsc{ApplyRule}$[R]$, we compute $
\Pr(a_t = \textsc{ApplyRule}[R] \mid a_{<t}, y)
= \mathsf{softmax}_{R}\left(g(\vect{h}_t)\right)$
where $g(\cdot)$ is a 2-layer MLP with a $\mathsf{tanh}$ non-linearity.
For \textsc{SelectColumn}, we compute
\begin{align*}
    &\tilde{\lambda}_{i} = \frac{\vect{h}_t W_Q^\text{sc} (y_i W_K^\text{sc})^T}{\sqrt{d_x}} \qquad
    \lambda_i = \softmax_i \bigl\{ \tilde{\lambda}_{i} \bigr\} \\
    &\Pr(a_t = \textsc{\small SelectColumn}[i] \mid a_{<t}, y) = \sum_{j=1}^{|y|} \lambda_{j} L^\text{col}_{j, i}
\end{align*}
and similarly for \textsc{SelectTable}.
We refer the reader to \citet{yinSyntacticNeuralModel2017a} for details.

\begin{figure}
    \centering
    \includegraphics[width=0.9\columnwidth]{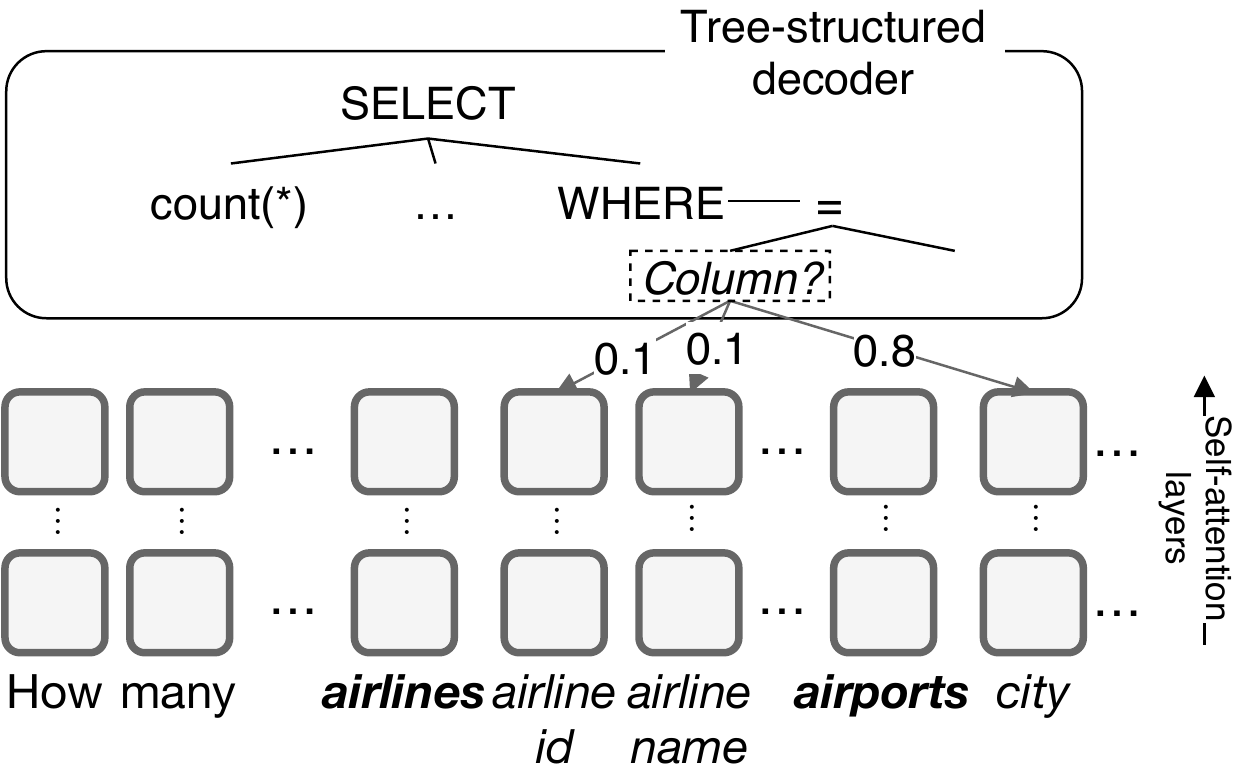}
    \caption{Choosing a column in a tree decoder.}
    \label{fig:decoder}
    \vspace{-\baselineskip}
\end{figure}

\section{Experiments}
\label{sec:experiments}

We implemented RAT-SQL in PyTorch~\citep{paszkeAutomaticDifferentiationPyTorch2017}.
During preprocessing, the input of questions, column names and table names are tokenized and lemmatized with the StandfordNLP toolkit~\citep{corenlp}.
Within the encoder, we use GloVe~\citep{pennington2014glove} word embeddings, held fixed in training except for the 50 most common words in the training set.
For RAT-SQL BERT, we use the WordPiece tokenization.
All word embeddings have dimension $300$.
The bidirectional LSTMs have hidden size 128 per direction, and use the recurrent dropout method of
\citet{galTheoreticallyGroundedApplication2016} with rate $0.2$.
We stack 8 relation-aware self-attention layers on top of the bidirectional LSTMs.
Within them, we set $d_x = d_z = 256$, $H = 8$, and use dropout with rate $0.1$.
The position-wise feed-forward network has inner layer dimension 1024.
Inside the decoder, we use rule embeddings of size $128$, node type embeddings of size $64$, and a hidden size of $512$
inside the LSTM with dropout of $0.21$.

We used the Adam optimizer \citep{kingmaAdamMethodStochastic2014} with the default hyperparameters.
During the first $warmup\_steps = max\_steps / 20$ steps of training, the learning rate linearly increases
from 0 to $\np{7.4e-4}$.
Afterwards, it is annealed to 0 with formula
$10^{-3}(1 - \frac{step - warmup\_steps}{max\_steps - warmup\_steps})^{-0.5}$.
We use a batch size of 20 and train for up to 40,000 steps.
For RAT-SQL + BERT, we use a separate learning rate of $\np{3e-6}$ to fine-tune BERT,
a batch size of 24 and train for up to 90,000 steps.

\paragraph{Hyperparameter Search}
We tuned the batch size (20, 50, 80), number of RAT layers (4, 6, 8), dropout (uniformly sampled from $[0.1, 0.3]$),
hidden size of decoder RNN (256, 512), max learning rate (log-uniformly sampled from $[\np{5e-4},\, \np{2e-3}]$).
We randomly sampled 100 configurations and optimized on the dev set.
RAT-SQL + BERT reuses most hyperparameters of RAT-SQL, only tuning the BERT learning rate (\np{1e-4},\, \np{3e-4},\, \np{5e-4}), number of RAT layers (6, 8, 10), number
of training steps (\np{4e4}, \np{6e4}, \np{9e4}).

\begin{table}
    \centering
    \small
    \begin{tabular}{lrr}
        \toprule
        \bfseries Model & \bfseries Dev & \bfseries Test \\
        \midrule
        IRNet \citep{irnet} & 53.2 & 46.7 \\
        Global-GNN \citep{bogin2019global} & 52.7 & 47.4 \\
        IRNet V2 \cite{irnet} & 55.4 & 48.5 \\
        \textbf{RAT-SQL} (ours) & \textbf{62.7} & \textbf{57.2} \\
        \midrule
        \textit{With BERT:} \\
        EditSQL + BERT \citep{EditSQL2019} & 57.6 & 53.4 \\
        GNN + Bertrand-DR \citep{bertrand} & 57.9 & 54.6 \\
        IRNet V2 + BERT \cite{irnet} & 63.9 & 55.0 \\
        RYANSQL V2 + BERT \citep{ryansql} & \textbf{70.6} & 60.6 \\
        \textbf{RAT-SQL + BERT} (ours) & 69.7 & \textbf{65.6} \\
        \bottomrule
    \end{tabular}
    \caption{Accuracy on the Spider development and test sets, compared to the other approaches at the top of the
        dataset leaderboard as of May 1st, 2020. The test set results were scored using the Spider evaluation server.}
    \label{table:main-leaderboard}
    \vspace{-1\baselineskip}
\end{table}

\subsection{Datasets and Metrics}
We use the Spider dataset \citep{data-spider} for most of our experiments, and also conduct preliminary experiments on
WikiSQL~\citep{zhongSeq2SQLGeneratingStructured2017}
to confirm generalization to other datasets.
As described by \citeauthor{data-spider}, Spider contains 8,659 examples (questions and SQL
queries, with the accompanying schemas), including 1,659 examples lifted from the Restaurants
\citep{data-restaurants-original,data-restaurants-logic}, GeoQuery \citep{data-geography-original}, Scholar
\citep{data-atis-geography-scholar}, Academic \citep{data-academic},
Yelp and IMDB \citep{data-sql-imdb-yelp} datasets.

As \citet{data-spider} make the test set accessible only through an evaluation server, we perform most evaluations (other than the final accuracy measurement) using the development set.
It contains 1,034 examples, with databases and schemas distinct from those in the training set.
We report results using the same metrics as \citet{yuSyntaxSQLNetSyntaxTree2018}:
exact match accuracy on all examples, as well as divided by difficulty levels.
As in previous work on Spider, these metrics do not measure the model's performance on generating values in the SQL.

\subsection{Spider Results}

In Table~\ref{table:main-leaderboard} we show accuracy on the (hidden) Spider test set for RAT-SQL and compare to all other approaches at or near state-of-the-art (according to the official leaderboard).
RAT-SQL outperforms all other methods that are not augmented with BERT embeddings by a large margin of 8.7\%.
Surprisingly, it even beats other BERT-augmented models.
When RAT-SQL is further augmented with BERT, it achieves the new state-of-the-art performance.
Compared with other BERT-argumented models, our RAT-SQL + BERT has smaller generalization gap between development and test set.

We also provide a breakdown of the accuracy by difficulty in Table~\ref{table:bydifficulty}.
As expected, performance drops with increasing difficulty.
The overall generalization gap between development and test of RAT-SQL was strongly affected by the significant drop in
accuracy (9\%) on the extra hard questions.
When RAT-SQL is augmented with BERT, the generalization gaps of 
most difficulties are reduced.

\begin{figure*}[t]
    \centering
    \includegraphics[width=0.95\textwidth]{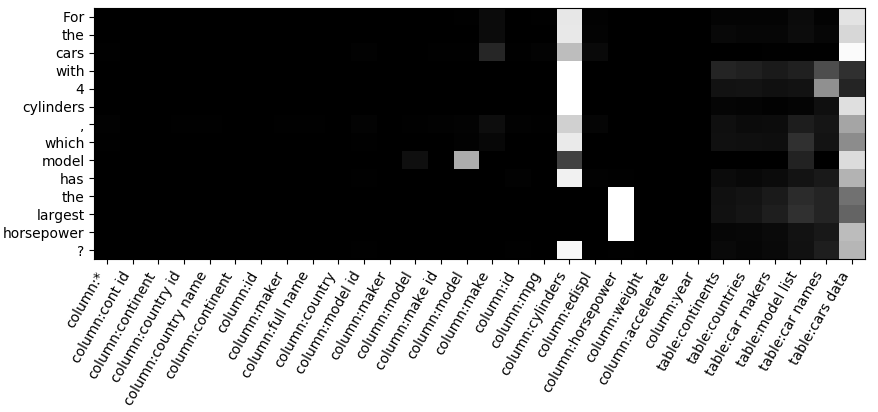}
    \vspace{-1\baselineskip}
    \caption{Alignment between the question \textit{``For the cars with 4 cylinders, which model has the largest horsepower''}
        and the database \texttt{car\_1} schema (columns and tables) depicted in \Cref{fig:intro:example}.}
    \label{fig:alignment}
    \vspace{-0.5\baselineskip}
\end{figure*}

\begin{table}
    \centering
    \small
    \begin{tabular}{lrrrrr}
        \toprule
        \bfseries Split  & \bfseries Easy & \bfseries Medium & \bfseries Hard & \bfseries Extra Hard & \bfseries All \\
        \midrule
        \multicolumn{5}{l}{\textit{RAT-SQL}} \\
        \bfseries Dev & 80.4 & 63.9 & 55.7 & 40.6 & 62.7 \\
        \bfseries Test &  74.8 & 60.7 & 53.6 & 31.5 & 57.2 \\
        \midrule
        \multicolumn{5}{l}{\textit{RAT-SQL + BERT}} \\
        \bfseries Dev & 86.4 & 73.6 & 62.1 & 42.9 & 69.7 \\
        \bfseries Test &  83.0 & 71.3 & 58.3 & 38.4 & 65.6 \\
        \bottomrule
    \end{tabular}
    \caption{Accuracy on the Spider development and test sets, by difficulty as defined by \citet{data-spider}.}
    \label{table:bydifficulty}
\end{table}

\begin{table}
    \centering
    \small
    \begin{tabular}{lr}
        \toprule
        \bfseries Model & \bfseries Accuracy (\%) \\
        \midrule
        \textbf{RAT-SQL + value-based linking} & \textbf{60.54 \textpm~0.80} \\
        RAT-SQL    & 55.13 \textpm~0.84 \\
        \quad w/o schema linking relations & 40.37 \textpm~2.32 \\
        \quad w/o schema graph relations   & 35.59 \textpm~0.85 \\
        \bottomrule
    \end{tabular}
    \vspace{-0.4mm}
    \caption{Accuracy (and \textpm 95\% confidence interval) of RAT-SQL ablations on the dev set.
        }
    \label{table:ablation-schemalinking}
    \vspace{-\baselineskip}
\end{table}

\paragraph{Ablation Study}
Table~\ref{table:ablation-schemalinking} shows an ablation study over different RAT-based relations.
The ablations are run on RAT-SQL without value-based linking to avoid interference with information from the database.
Schema linking and graph relations make statistically significant improvements (p<0.001).
The full model accuracy here slightly differs from Table~\ref{table:main-leaderboard} because
the latter shows the best model from a hyper-parameter sweep (used for test evaluation) and the former gives
the mean over five runs where we only change the random seeds.

\subsection{WikiSQL Results}
\label{sec:wikisql}

\begin{table*}
    \centering
    \small
    \begin{tabular}{lrrrr}
        \toprule
        & \multicolumn{2}{c}{\bfseries Dev} & \multicolumn{2}{c}{\bfseries Test} \\
        \cmidrule{2-5}
        \bfseries Model & \bfseries LF Acc\% & \bfseries Ex. Acc\% & \bfseries LF Acc\% & \bfseries Ex. Acc\% \\
        \midrule
        IncSQL~\citep{shiIncSQLTrainingIncremental2018} & 49.9 & 84.0 & 49.9 & 83.7 \\
        MQAN~\citep{mccann2018natural} & 76.1 & 82.0 & 75.4 & 81.4 \\
        RAT-SQL (ours) & 73.6 & 79.5 & 73.3 & 78.8 \\
        Coarse2Fine~\citep{dongCoarsetoFineDecodingNeural2018} & 72.5 & 79.0 & 71.7 & 78.5 \\
        PT-MAML~\citep{huang2018natural} & 63.1 & 68.3 & 62.8 & 68.0 \\
        \bottomrule
    \end{tabular}
    \caption{
        RAT-SQL accuracy on WikiSQL, trained without BERT augmentation or execution-guided decoding~(EG).
        Compared to other approaches without EG.
        ``LF Acc'' = Logical Form Accuracy; ``Ex. Acc'' = Execution Accuracy.
    }
    \label{tab:wikisql}
    \vspace{-\baselineskip}
\end{table*}

We also conducted preliminary experiments on WikiSQL~\citep{zhongSeq2SQLGeneratingStructured2017} to test
generalization of RAT-SQL to new datasets.
Although \mbox{WikiSQL} lacks multi-table schemas (and thus, its challenge of schema encoding is not as prominent), it still
presents the challenges of schema linking and generalization to new schemas.
For simplicity of experiments, we did not implement either BERT augmentation or execution-guided decoding~(EG)~\citep{wangRobustTexttoSQLGeneration2018},
both of which are common in state-of-the-art \mbox{WikiSQL} models.
We thus only compare to the models that also lack these two enhancements.

While not reaching state of the art, RAT-SQL still achieves competitive performance on WikiSQL as shown in
\Cref{tab:wikisql}.
Most of the gap between its accuracy and state of the art is due to the simplified implementation of \emph{value
decoding}, which is required for WikiSQL evaluation but not in Spider.
Our value decoding for these experiments is a simple token-based pointer mechanism, which often fails to retrieve
multi-token value constants accurately.
A robust value decoding mechanism in RAT-SQL is an important extension that we plan to address outside
the scope of this work.

\subsection{Discussions}

\paragraph{Alignment}

Recall from Section~\ref{sec:our-encoder} that we explicitly model the alignment matrix between question words and table
columns, used during decoding for column and table selection.
The existence of the alignment matrix provides a mechanism for the model to align words to columns. 
An accurate alignment representation has other benefits such as identifying question words to copy to emit a constant value in SQL.

In Figure~\ref{fig:alignment} we show the alignment generated by our model on the example from \Cref{fig:intro:example}.\footnote{
The full alignment also maps from column and table names, but those end up simply aligning to themselves or the table
they belong to, so we omit them for brevity.}
For the three words that reference columns (\textit{``cylinders''}, \textit{``model''}, \textit{``horsepower''}), the alignment matrix
correctly identifies their corresponding columns.
The alignments of other words are strongly affected by these three keywords, resulting in a sparse span-to-column like alignment, \eg \textit{``largest horsepower''} to \texttt{horsepower}.
The tables \mbox{\texttt{cars\_data}} and \mbox{\texttt{cars\_names}}
are implicitly mentioned by the word \textit{``cars''}.
The alignment matrix successfully infers to use these two tables instead of \mbox{\texttt{car\_makers}} using the evidence that they contain the three mentioned columns.

\paragraph{The Need for Schema Linking}
One natural question is how often does the decoder fail to select the correct column, even with the schema encoding and linking improvements we have made.
To answer this, we conducted an oracle experiment (see Table~\ref{table:oracle}).
For ``oracle sketch'', at every grammar nonterminal the decoder is forced to choose the correct production so the final SQL
sketch exactly matches that of the ground truth.
The rest of the decoding proceeds conditioned on that choice.
Likewise, ``oracle columns'' forces the decoder to emit the correct column/table at terminal nodes.

With both oracles, we see an accuracy of 99.4\% which just verifies that our grammar is sufficient to answer nearly
every question in the data set.
With just ``oracle sketch'', the accuracy is only 73.0\%, which means 72.4\% of the questions that RAT-SQL gets wrong
and could get right have incorrect column or table selection.
Similarly, with just ``oracle columns'', the accuracy is 69.8\%, which means that 81.0\% of the questions that RAT-SQL gets
wrong have incorrect structure.
In other words, most questions have both column and structure wrong, so both problems require important future work.

\begin{table}[t]
    \centering
    \small
    \begin{tabular}{lr}
        \toprule
        \bfseries Model & \bfseries Acc. \\
        \midrule
        RAT-SQL & 62.7 \\
        RAT-SQL + Oracle columns & 69.8 \\
        RAT-SQL + Oracle sketch & 73.0 \\
        RAT-SQL + Oracle sketch + Oracle columns & 99.4 \\
        \bottomrule
    \end{tabular}
    \caption{Accuracy (exact match \%) on the development set given an oracle providing correct columns and tables
        (``Oracle columns'') and/or the AST sketch structure (``Oracle sketch'').
    }
    \vspace{-\baselineskip}
    \label{table:oracle}
\end{table}

\paragraph{Error Analysis}
An analysis of mispredicted SQL queries in the Spider dev set showed three main causes of evaluation errors.
\textbf{(I)} 18\% of the mispredicted queries are in fact \emph{equivalent} implementations of the NL intent with a different SQL syntax (\eg \texttt{ORDER BY $C$ LIMIT 1} vs. \texttt{SELECT MIN($C$)}).
Measuring execution accuracy rather than exact match would detect them as valid.
\textbf{(II)} 39\% of errors involve a wrong, missing, or extraneous column in the \texttt{SELECT} clause.
This is a limitation of our schema linking mechanism, which, while substantially improving column resolution, still struggles with some ambiguous references.
Some of them are unavoidable as Spider questions do not always specify which columns should be returned by the desired SQL.
Finally, \textbf{(III)}~29\% of errors are missing a \texttt{WHERE} clause, which is a common error class in text-to-SQL models as reported by prior works.
One common example is domain-specific phrasing such as \textit{``older than 21''}, which requires background knowledge to map it to \texttt{age > 21} rather than \texttt{age < 21}.
Such errors disappear after in-domain fine-tuning.

\section{Conclusion}
\label{sec:conclusion}

Despite active research in text-to-SQL parsing, many contemporary models struggle to learn good
representations for a given database schema as well as to properly link column/table references in the question.
These problems are related: to encode \& use columns/tables from the schema, the model must reason about their role in
the context of the question.
In this work, we present a unified framework for addressing the schema encoding and linking challenges.
Thanks to relation-aware self-attention, it jointly learns schema and question representations based on their
alignment with each other and schema relations.

Empirically, the RAT framework allows us to gain significant state of the art improvement on text-to-SQL parsing.
Qualitatively, it provides a way to combine predefined \emph{hard} schema relations and inferred \emph{soft}
self-attended relations in the same encoder architecture.
This representation learning will be beneficial in tasks beyond text\hyp{}to\hyp{}SQL, as long as the
input has some predefined structure.

\section*{Acknowledgments}
We thank Jianfeng Gao, Vladlen Koltun, Chris Meek, and Vignesh Shiv for the discussions that helped shape this work.
We thank Bo Pang, Tao Yu for their help with the evaluation.
We also thank anonymous reviewers for their invaluable feedback.

\bibliography{text2sql,text2sql-datasets}
\bibliographystyle{acl_natbib}

\newpage
\appendix

\section{Auxiliary Relations for Schema Encoding}
\label{app:schema-encoding}
In addition to the schema graph edges $\mathcal{E}$ (\Cref{sec:rel-attn})
and schema linking edges (\Cref{sec:schema-linking}),
the edges in $ \mathcal{E}_{\question} $ also include some auxiliary relation types
to aid the relation-aware self-attention.
Specifically, for each $x_i, x_j \in \mathcal{V}_{\question}$:
\begin{itemize}
    \item If $i = j$, then \textsc{Column-Identity} or \textsc{Table-Identity}.
    \item $x_i \in \question$, $x_j \in \question$:
       \textsc{Question-Dist-$d$}, where
       \begin{align*}
       d &= \mathsf{clip}(j - i, D), \\
       \mathsf{clip}(a, D) &= \max(-D, \min(D, a)).
       \end{align*}
       We use $D = 2$.
    \item Otherwise, one of \textsc{Column-Column}, \textsc{Column-Table}, \textsc{Table-Column}, or
        \textsc{Table-Table}.
\end{itemize}

\section{Alignment Loss}
\label{app:align_loss}

The memory-schema alignment matrix is expected to resemble the real discrete alignments, therefore
should respect certain constraints like sparsity.
For example, the question word \textit{``model''} in \Cref{fig:intro:example} should be aligned with
\texttt{car\_names.model} rather than \mbox{\texttt{model\_list.model}} or \mbox{\texttt{model\_list.model\_id}}.
To further bias the soft alignment towards the real discrete structures,
we add an auxiliary loss to encourage sparsity of the alignment matrix.
Specifically, for a column/table that is mentioned in the SQL query,
we treat the model's current belief of the best alignment as the ground truth.
Then we use a cross-entropy loss, referred as \textit{alignment loss},
to strengthen the model's belief:
\begin{align*}
    align\_loss = &- \frac{1}{|Rel(\sColumnSet)|} \sum_{j \in Rel(\sColumnSet)} \log \max_i L^{\text{col}}_{i,j} \\
    &- \frac{1}{|Rel(\sTableSet)|} \sum_{j \in Rel(\sTableSet)} \log \max_i L^{\text{tab}}_{i,j}
\end{align*}
where $Rel(\sColumnSet)$ and $Rel(\sTableSet)$ denote the set of \emph{relevant} columns and tables that appear in the SQL.

In earlier experiments, we found that the alignment loss did improve the model (statistically significantly, from 53.0\% to 55.4\%). 
However, it does not make a statistically significant difference in our final model in terms of overall exact match.
We hypothesize that hyperparameter tuning that caused us to increase encoding depth eliminated the need for explicit supervision of alignment.
With few layers in the Transformer, the alignment matrix provided additional degrees of freedom, which became unnecessary once the Transformer was sufficiently deep to build a rich joint representation of the question and the schema.

\section{Consistency of RAT-SQL}
\label{app:consistency}

In Spider dataset, most SQL queries correspond to more than one question, making it possible to evaluate the consistency of RAT-SQL given paraphrases. We use two metrics to evaluate the consistency: 1) \textit{Exact Match} -- whether RAT-SQL produces the exact same predictions given paraphrases, 2) \textit{Correctness} -- whether RAT-SQL achieves the same correctness given paraphrases. The analysis is conducted on the development set.

\begin{table}[t]
\centering
\small
\begin{tabular}{ lcc }
  \toprule
\textbf{Model} & \textbf{Exact Match} & \textbf{Correctness} \\
  \midrule
\textbf{RAT-SQL}  & 0.59 & 0.81 \\
\textbf{RAT-SQL + BERT} & 0.67 & 0.86 \\
\bottomrule
\end{tabular}
\caption{Consistency of the two RAT-SQL models.}
\label{tab:consistency}
\end{table}

The results are shown in \Cref{tab:consistency}. We found that when augmented with BERT, RAT-SQL becomes more consistent in terms of both metrics, indicating the pre-trained representations of BERT are beneficial for handling paraphrases.

\end{document}